# Optimizing Deep Neural Network Architecture: A Tabu Search Based Approach

Tarun Kumar Gupta and Khalid Raza*

Department of Computer Science, Jamia Millia Islamia, New Delhi-110025
`*kraza@jmi.ac.in`

**Abstract.** The performance of Feedforward neural network (FNN) fully depends upon the selection of architecture and training algorithm. FNN architecture can be tweaked using several parameters, such as the number of hidden layers, number of hidden neurons at each hidden layer and number of connections between layers. There may be exponential combinations for these architectural attributes which may be unmanageable manually, so it requires an algorithm which can automatically design an optimal architecture with high generalization ability. Numerous optimization algorithms have been utilized for FNN architecture determination. This paper proposes a new methodology which can work on the estimation of hidden layers and their respective neurons for FNN. This work combines the advantages of Tabu search (TS) and Gradient descent with momentum backpropagation (GDM) training algorithm to demonstrate how Tabu search can automatically select the best architecture from the populated architectures based on minimum testing error criteria. The proposed approach has been tested on four classification benchmark dataset of different size

## 1    Introduction

Artificial neural network (ANN) is a mathematical model which was designed to mimic the way the human brain processes information. They have an input layer, an output layer and one or more hidden layers. These hidden layers act as computational engine for the network. Its simple design, good learning ability and capability of solving complex problems make ANN very popular.

In ANN model, the number of processing units (neurons) at input and output layer are fixed while these vary at the hidden layers. There is no straightforward criterion to calculate number of neurons in hidden layers nor is there any supporting theory for calculating the hidden layers. These architectural attributes are involved in the performance of ANN because a network with few layers and few neurons can cause under-fitting while a large network easily leads to overfitting. Also, ANN with different structure gives different output for the same data set. Therefore, architectural design of ANN is very crucial and may be defined as an optimization problem [1]. In ANN optimization, populated architectures are called solutions and their testing error may be considered as a cost function. Hence, the challenge is to find an optimum architecture with the lowest testing error using some optimization techniques.

Usually, the selection of ANN architecture is based on hit & trial principle, which is time-consuming and poses many challenges, for example- initializing number of hidden layers, hidden neurons, connections, etc. requires pre-knowledge about the ANN functioning and problem domain. This initialization of parameters cannot be accomplished without any expertise because there may be exponential combinations for these attributes. So, the process of selecting parameters by hit & trial method needs much human intensive hard work without any guarantee of obtaining an accurate model. Moreover, when the problem is from a very complex domain then deciding parameters for ANN becomes quite a difficult and tedious process. There are many optimization algorithms which can be utilized to manage these issues; including genetic algorithm (GA) [2], simulated annealing (SA) [3], tabu search (TS) [4] and bat algorithm [5].

This paper proposes a new methodology which combines the advantages of Tabu search and Gradient descent with momentum backpropagation training algorithm [6] for finding the optimal design for a feedforwars neural network. It automatically handles the problem of defining hidden layers and their respective neurons, which in earlier cases used to be a manual task. The application of this methodology is tested over four different classification data sets: 1) the face recognition dataset [7] 2) Gas Senser Array Drift dataset [8], [9] 3) MNIST dataset of handwritten digits [10] and 4) ISOLET dataset [11].

The paper is structured as follows: Section 2 describes related works on feedforward neural network (FNN) optimization using different optimization techniques. Section 3 presents optimization methodology used here with its components like solution representation, fitness function, population generation and stopping mechanism. Section 4 describes each data sets with their properties used in the study. Section 5 present experimental setup and results. Finally, in Section 6, the paper covers discussion and future scope.

## 2    Related Work

Over the last two decades, plenty of algorithms have been proposed for optimizing training rule as well as ANN architecture which can suggest an optimal or nearest optimal ANN structure but problem still needs further exploration. For example, Stepniewski and Keane [12] applied an integration of GA and SA based pruning algorithm as a stochastic optimization technique. GA translated the ANN architecture into a chromosome sequence and SA was used to find the optimum architecture from translated sequences. Initially, overdetermined multilayer perceptron (MLP) architecture was chosen as a starting point and then it was made to shrink by the elimination of useless links and nodes. Ludermir et al. [13] proposed TS and SA based algorithm to optimize MLP i.e., weight and architecture simultaneously. In this work SA suggests an approach to determine whether to accept the new solution or not on the basis of probability, whereas TS evaluates a batch of solutions in single iteration which avoids low convergence. Geppert and Roth [14] employed a multi-objective evolutionary process to optimize ANN architecture. The process integrates SA with TS and applied to two real-world problems car classification and face recognition. Tsai et al. [15] used a hybrid Taguchi-genetic algorithm (HTGA) to help to design the parameter of FNN. A particle swarm optimization (PSO) based ANN design was proposed in [16]. However, in the same year, an improved version of PSO and discrete PSO [17] evaluated three-layer FNN architecture and parameters.

Islam et al. [18], [19] present the use of adaptive merging and growing algorithm (AMGA) for designing three-layer FNN. Initially the hidden neurons were selected randomly and then AMGA pruning process merged or added neurons on the basis of their learning ability, while [20] proposed a hybrid multiobjective evolutionary algorithm to measure the minimum requirement of hidden neurons for FNN with single hidden layer. In [21], authors applied evolutionary algorithm with multilogistic regression to design architecture and weight of ANN. Mantzaris et al. [22] employed Probabilistic Neural Networks (PNNs) to prune ANN architecture using a genetic algorithm. This pruning reduces the size of input and hidden layer.

In [23], authors proposes GaTSa (combination of simulated annealing, tabu search, and genetic algorithm). In this approach, the algorithm chose initial population with a random number of neurons at hidden layer and categorized them by rank-based fitness scaling [24], then GA updated hidden nodes in a constructive way using universal stochastic sampling for next iteration. Here, SA helps in moving out from local minima, and TS evaluates a set of the solutions in single iteration leading to faster convergence. In another work, Jaddi et al. [25] proposed modified bat algorithm to search optimized ANN structure and tuned it with Taguchi method [26]. Later on, the author did the same experiments using multi population based cooperative bat algorithm [27]. Further, Jaddi [28] proposed a new algorithm for solution representation, genetic algorithm based dynamic neural network (GADNN). The solution consists of two vectors, in first vector half part contains information of hidden layer and other half part represents hidden neurons, while as the second vector stores weights and biases. Recent review in this area can be found in [29].

From the literature we can conclude that there exist some important lacunae in FNN architecture optimization techniques, such as a) most of the algorithms [12]–[16], [20], [21] require predefined size for chromosomes; it influences the performance of the algorithm because the manual definition of chromosome size is problem dependent, b) merging and pruning techniques [18], [19] require much attention in pre-defining rules about when to remove or add hidden neuron, and c) almost every algorithm works only for single hidden layer; less consideration has been given towards the optimization of deep neural networks.

## 3 Optimization methodology

Tabu search (TS) is a meta-heuristic approach that has been acknowledged to be very effective and has a broad consideration in diverse problem domains because of its adaptability and many considerable achievements in finding globally optimal solu-tions. The strategy for neighborhood search and the use of long and short-term memories distinguishes tabu search from local search and other heuristic methods. Tabu search method can calculate a batch of solutions in a single iteration which minimizes computational cost and tends to faster convergence. The best solution (with the lowest cost) of the latest iteration can be accepted as the current solution for the next iteration. The strategy maintains a list (tabu list) which records the last visited solutions to avoid the repetition.

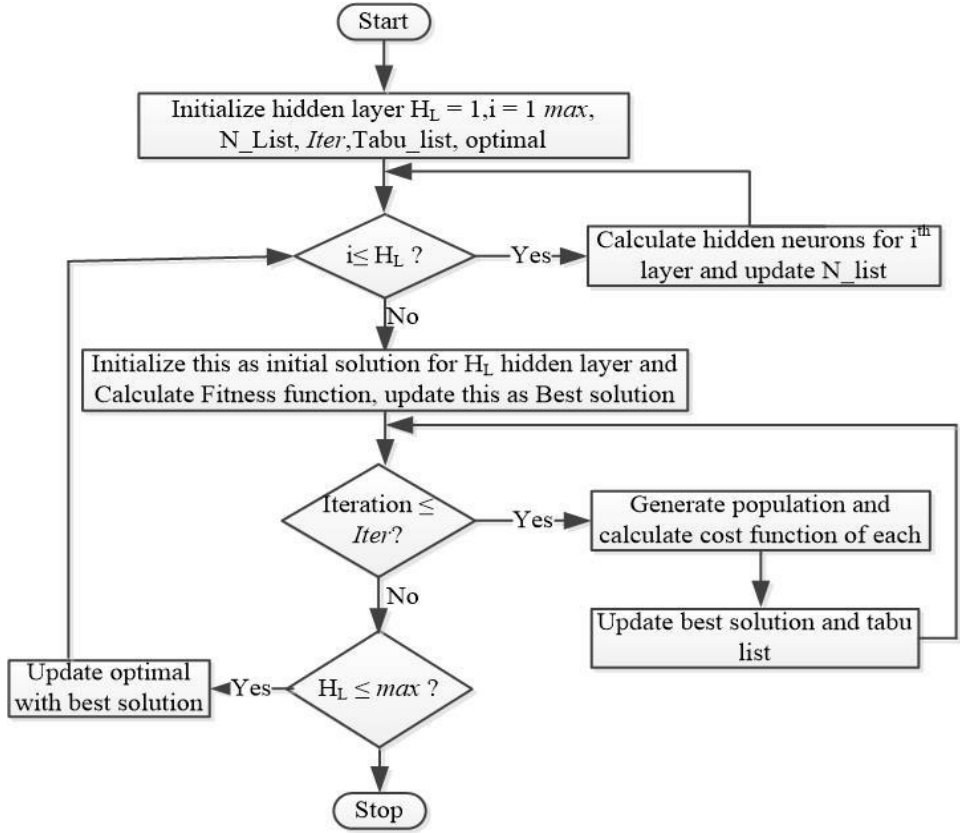

**Fig. 1.** Flowchart of the Proposed algorithm

In this paper, Tabu search is integrated with Gradient descent with momentum backpropagation (GDM). The aim of this optimization is to search solution $s$ from the set of solutions $S$, such that $f(s) \leq f(s')$, for all $s' \in S$. Methodology starts with FNN having one hidden layer and randomly selected hidden neurons. Figure 1 shows the flow chart of proposed methodology. The proposed methodology runs for maximum $max$ hidden layers. A solution gets iterated $Iter$ times, each iteration generates a population of size $P_{max}$, and TS chooses best on the basis of the fitness function. If the best of $Iter$ called $s'$ is less than $s_{best}$ then update $s_{best}$ and go to next iteration; otherwise update Tabu list and further explore $s'$ until no stopping criteria is met.

The pseudocode of proposed methodology is shown in Algorithm 1. For implementation of proposed strategy following definitions need to be considered: 1) solution representation; 2) the fitness function; 3) population generation, and 4) stopping mechanism.

---

**ALGORITHM 1:** Pseudocode for Proposed Methodology

---

**INPUT:** $\# input\ neurons, \# output\ neurons, max, Iter$

$For\ H_L\ =\ 1\ to\ max$
{
  $Input \leftarrow \#input\ neurons$
  $Output \leftarrow \#\ output\ neurons$
  $N\_List[\ ] \leftarrow NULL$
  $Tabu\_List[\ ]\ \leftarrow NULL$
  $For\ H_N\ =\ 1\ to\ H_L$
  {
    $a \leftarrow \frac{(Input+Output)}{2}$
    $b \leftarrow \frac{(Input+Output)\times 2}{3}$
    $N\_List[H_N]\ \leftarrow\ random(a,b)$
    $Input \leftarrow N\_List[H_N]$
  }
  $s_0 \leftarrow calculate\_fitness(H_L,\ N\_List)$
  $s_0\ is\ the\ initial\ solution\ update\ with\ s_{best}$
  $Tabu\_List[\ ] \leftarrow s_{best}$
  $For\ x\ =\ 1\ to\ Iter$
  {
    $s' \leftarrow Generate\_Neighbor(s_{x-1})$
    $s'\ is\ best\ from\ neighbor\ of\ s_{x-1}$
    $if\ f(s')\ <\ f(s_{best})$
    {
      $s_{best} \leftarrow s'$
      $s_x \leftarrow s'$
    }
    $else$
    {
      $Tabu\_List[next] \leftarrow s'$
      $s_x \leftarrow s'$
    }
  }
  $optimal[H_L] \leftarrow s_{best}$   //list contains best architecture
}
$Return\ best\ of\ optimal[H_L]$

---

### 3.1 Solution Representation

FNN with multiple hidden layers is considered for evaluation. Each network contains $I$ input nodes, $H_i$ hidden nodes at '$i^{th}$' hidden layer and $O$ output nodes. Generally, the number of input and output nodes is problem specific; the aim is to find out the optimal number of hidden layers and their respective nodes. Architecture of a FNN can be written as:

$$A \equiv (I \times H_1 + B \times H_1) + (H_1 \times H_2 + B \times H_2) + \cdots + (H_{max} \times O + B \times O) \quad (1)$$

Where 'B' is bias.

Every solution consist of three attributes: $H_L$, containing the number of hidden layers, $H_N$, a vector containing the number of hidden neurons (layer-wise), $T_E$ represents the testing error of the given solution,

$$S \equiv (H_L, H_N, T_E) \quad (2)$$

$$H_N \equiv (H_1, H_2, H_3, \ldots, H_{max}), H_i \in \mathbb{N}, \forall H_{(i-1)} > H_i > H_{(i+1)} \text{ and } i = 1,2, \ldots, max \quad (3)$$

$$H_L \equiv (1,2,3,\ldots,\text{max}) \text{ and } T_E \in \Re \quad (4)$$

Where $\mathbb{N}$ is set of natural numbers and $\Re$ is a set of real numbers

The initial solution takes a fully connected FNN containing a fixed number of input and output neurons with a single hidden layer. Hidden neurons for the layers are calculated by random selection in between the range of [(I+O)/2, (I+O)×2/3] and the initial weights are in between the interval of [-1.0, +1.0] uniformly distributed.

### 3.2 Fitness Function

A fitness function is a percentage of the accuracy of a given model in terms of its capability during the approximation of results. It requires comparing the performance of solution in successive iterations to select a solution that minimizes the objective function.

Let the given dataset be divided into $C_N$ classes, and the actual class of sample $d$ from testing set $T$ be written as:

$$\lambda(d) \in \{1,2,3,\ldots,C_N\} \forall d \in T \quad (5)$$

The proposed technique uses *winner takes all* method, so the number of output neurons and the classes $C_N$ of given dataset has one to one correspondence, $O_p(d)$ being the value from output node $p$ for sample $d$, the class for the sample $d$ is:

$$\Omega(d) \equiv arg\ max_{p\ \epsilon\ \{1,2,3,\ldots,C_N\}} O_p(d) \qquad \forall d\ \epsilon\ T \quad (6)$$

The error for the sample d is:

$$\varepsilon(d) \equiv \begin{cases} 1, & if\ \lambda(d) \neq \Omega(d) \\ 0, & if\ \lambda(d) = \Omega(d) \end{cases} \quad (7)$$

Hence, the classification error of network for testing set $T$, i.e., misclassified samples during the testing phase in terms of percentage is written as:

$$E(T) \equiv \frac{100}{T} \sum_{d \in T} \varepsilon(d) \qquad (8)$$

Where $\#T$ is the cardinality of testing set $T$.

---

**ALGORITHM 2:** Pseudocode for *Generate_Neighbor*

---

**INPUT**: $P_{max}$, $p$, $K$
$Candidate\_List[P_{max}] \leftarrow NULL$
*For* $i = 1$ to $(P_{max}/2)$
{
        $Ft \leftarrow NULL$
    *For* $j = 1$ to $H_L$
    {
        $\omega = random(0, 1)$
        *if* $\omega \geq p$      // $p$ is probability
    Increase number of neurons by '$K$' at that layer
        *else*
            No change with neurons at that layer
    }
    Update $N\_List[\ ]$ for $Candidate\_List[i]$
    $Ft \leftarrow calculate\_fitness(Candidate\_List[i])$
        Update $Ft$ of $Candidate\_List[i]$
}
*For* $i = 1$ to $(P_{max}/2)$
{
        $Ft \leftarrow NULL$

    *For* $j = 1$ to $H_L$
    {
        $\omega = random(0, 1)$
        *if* $\omega \geq p$    // $p$ is probability
            decrease number of neurons by '$K$' at that layer
        *else*
            No change with neurons at that layer
    }
    Update $N\_List[\ ]$ for $Candidate\_List[i + P_{max}/2]$
    $Ft \leftarrow calculate\_fitness(Candidate\_List[i + P_{max}/2])$
    Update $Ft$ of $Candidate\_List[i + P_{max}/2]$
}
Return best of $candidate\_List[P_{max}]$

---

### 3.3    Population Generation

After evaluation of initial solution $s = (H_L, H_N, T_E)$, the proposed methodology populates a complete generation of solutions using algorithm 2 until stopping criteria is met. Each solution of new population is generated in the following manner: divide the size of the population in two equal parts, one for increasing and other for decreasing neurons at a particular layer. Case 1, generate a random number $\omega$ uniformly distributed in [0, 1] for each layer of every new solution,

$$H_N = \begin{cases} K+, & \omega \geq p \\ no\ change, & \omega < p \end{cases} \tag{9}$$

Where $K +$ signifies an increase in the number of neurons by percentage $K$ until the value does not exceed the upper boundary. Similarly, Case 2; generates a random number $\omega$ such that:

$$H_N = \begin{cases} K-, & \omega \geq p \\ no\ change, & \omega < p \end{cases} \tag{10}$$

Where $K -$ signifies the reducing of number of neurons by $K$ percent, until lower limit is reached. This process of generating new solutions makes it possible to move in both forward and backward directions in order to search the global optimal solution.

### 3.4    Stopping mechanism

In this subsection, the algorithm will stop after optimizing $max$ hidden layers up to $Iter$ iterations. However, in some cases it avoids updating neurons and jumps to successive step: 1) when algorithm tries to increase the number of neurons and it already reaches to its upper limit, and 2) when algorithm tries to decrease neurons and it already reaches to its lower limit.

## 4    Datasets

We used four different classification datasets as presented in Table 1, to validate the proposed algorithm. According to the problem we require datasets having a large number of features because with few features, algorithm would not show its efficiency and will converge with a single hidden layer.

### 4.1    Face recognition dataset

In this, the dataset consists of high-resolution male and female images between the ages of 18 to 40 years with different emotions; taken from the *CHICAGO FACE DATABASE (CFD)* developed at the University of Chicago. Every image is in *JPEG* format. There is a total of 1846 images where 972 are males and 874 are females. The aim is to classify these images into two classes i.e., male and female. To do this the images are converted from JPEG format to vector format, every vector is of dimension 1x785. From column 1 to 784 every vector describes features of the image and the last column defines the label.

### 4.2 Gas Senser Array Drift dataset

This dataset holds 13910 examples from 16 chemical sensors employed in recreations for drift compensation in a classification task of 6 gases at different levels of concentrations.The samples were collected from a gas delivery platform (Jan' 2007 to Feb' 2011) in a fully computerized environment for minimizing the common errors caused by human involvement. The purpose is to accomplish good performance (or as low degradation as possible) over time. The dataset has 128 inputs and 6 outputs.

### 4.3 MNIST dataset of handwritten digits

The MNIST (Modified National Institute of Standards and Technology) database is a huge database containing handwritten digits that are generally used for training many image processing systems. The database was generated from the samples of NIST's datasets. Every image is in bi-level format and normalized in 28x28 pixel box. The MNIST dataset contains 70,000 samples, with 784 inputs and 10 outputs. In this problem, the aim is to classify every image in 10 distinct classes from 0 to 9.

### 4.4 ISOLET dataset

ISOLET (Isolated Letter Speech Recognition) dataset was created in the following manner: there are 150 speakers who spoke each alphabet twice. The author divides 150 speakers into 5 groups, hence each group is of 30 speakers, and the dataset from 4 groups was selected for training purpose and one group for testing. The recorded sample has 617 features and needs to be classified into 26 distinct classes. The dataset contains a total of 7797 samples.

**Table 1.** Dataset Statistics

| Dataset | Examples | Features | Classes | References |
|---------|----------|----------|---------|------------|
| Face | 1846 | 784 | 2 | [7] |
| Gas-Drift | 13910 | 128 | 6 | [8],[9] |
| MNIST | 70000 | 784 | 10 | [10] |
| ISOLET | 7797 | 617 | 26 | [11] |

## 5 Experimental Setup and Results

In the experiment, proposed algorithm is implemented using R. Every network architecture is validated by randomly selected (20% of dataset) validation set. The imple-mentation uses rectifier with dropout activation function where input dropout ratio is taken as 0.2. The datasets are normalized by min-max method.

The effectiveness of proposed methodology is tested over four classification datasets (Table 1). The main contribution of this paper is to find the optimal architecture of deep FNN when the dataset has a large number of features and requires more than one hidden layers. In our experiment except for face data, rest all datasets are of more than 600 features and multiple classes

**Table 2.** Experimental results by Proposed method

| Dataset | Hidden Layer | Hidden Neurons | Training Error | Testing Error |
|---|---|---|---|---|
| **Face recognition** | 1 | 515 | 2.6125 | 11.396 |
| | **2** | **437,260** | 8.0481 | **10.3859** |
| | 3 | 506,261,140 | 9.5684 | 11.4385 |
| | 4 | 454,261,148,73 | 8.4915 | 11.0072 |
| | 5 | 445,250,156,189,48 | 11.6518 | 13.176 |
| **Gas-Drift** | 1 | 75 | 5.3171 | 6.3471 |
| | **2** | **90,62** | 4.9861 | **5.02398** |
| | 3 | 78,53,36 | 6.0466 | 6.2952 |
| | 4 | 74,79,35,25 | 10.5086 | 10.7071 |
| | 5 | 79,45,33,25,19 | 25.4556 | 25.7846 |
| **MNIST** | **1** | **518** | 0.651 | **1.823** |
| | 2 | 547,225 | 0.645 | 1.902 |
| | 3 | 520,289,165 | 1.1113 | 1.926 |
| | 4 | 532,302,130,75 | 1.2895 | 2.057 |
| | 5 | 490,274,158,103,64 | 3.2325 | 3.081 |
| **ISOLET** | 1 | 335 | 0.0301 | 2.0698 |
| | **2** | **362,231** | 0.2726 | **1.8297** |
| | 3 | 397,212,119 | 0.8519 | 2.3935 |
| | 4 | 406,254,155,100 | 3.3127 | 4.6555 |
| | 5 | 402,227,135,78,46 | 92.448 | 91.484 |

The proposed methodology starts with layer one and some randomly selected neurons for this layer. After calculation of fitness function, this initial solution can be thought as the best solution and all the searching for the global optimal solution starts from here. Every selected solution is iterated by $iter = 10$ and in each iteration $P_{max} = 20$. Futher the $P_{max}$ is splited into two parts, where one part increases the number of neurons at selected layer and other part decreases the same. Here, the value of $p$ (probability of changing neurons at particular layer) is set to 0.5 and the updation of neurons ($K$) is set to 3%. The methodology runs for $H_{max}$=5 layers, which can be increased if needed.

The results for proposed methodology are shown in Table 2. For instance, in face dataset, the optimized architecture according to the proposed approach was $H_L = 2$ and $H_N = \{437,260\}$ with mean classification error 10.38 % . In the gas drift dataset, the best topology with $H_L = 2$ and $H_N = \{90,62\}$ and the mean classification error was of around 5.023%. In MNIST dataset, the mean classification error was 1.823% and optimal topology with $H_L = 1$ and $H_N = 518$. For the isolet dataset, the optimal architecture requires $H_L = 2$ and $H_N = \{362,231\}$, a mean classification error was noted as 1.829%. The performance of proposed algorithm in terms of training and testing error for different $H_L = \{1,2,3,4,5\}$ can be seen in Fig.2

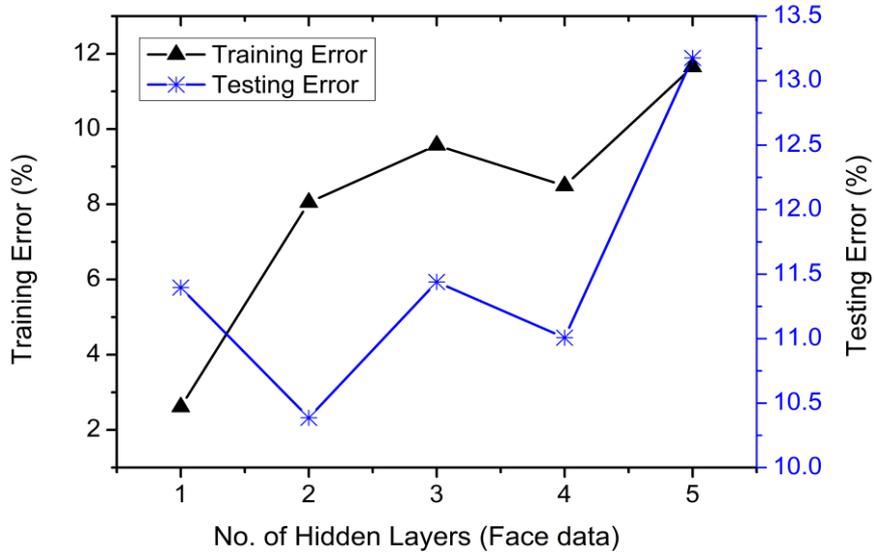

**a.** Performance of proposed algorithm on Face recognition dataset

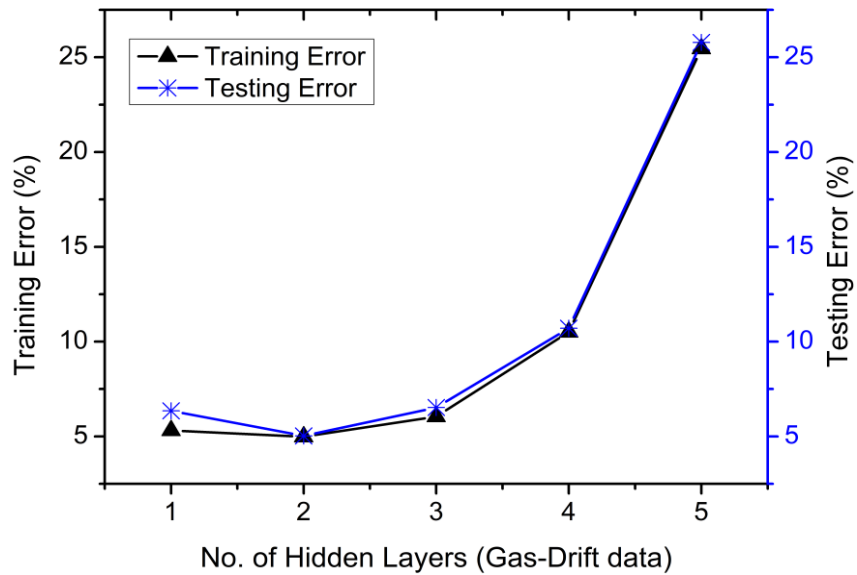

**b.** Performance of proposed algorithm on Gas-Drift dataset

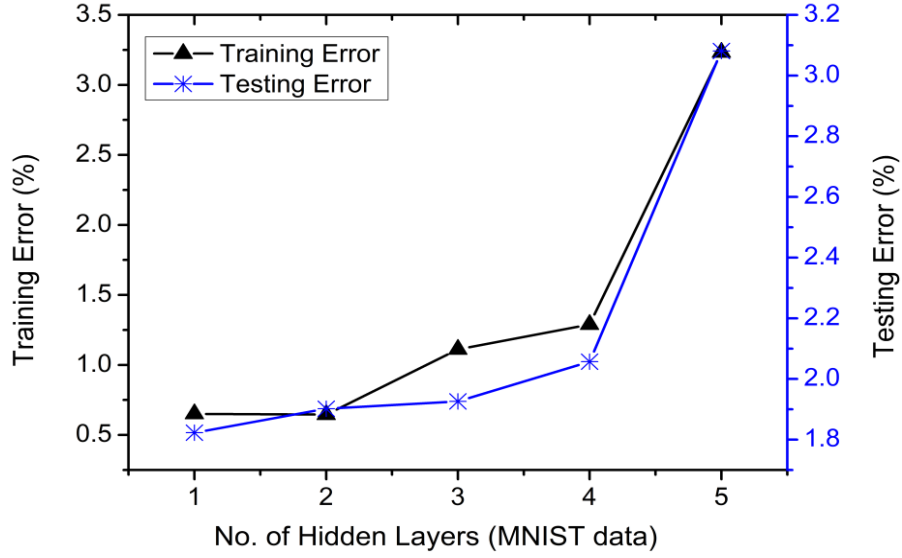

**c.** Performance of proposed algorithm on MNIST dataset

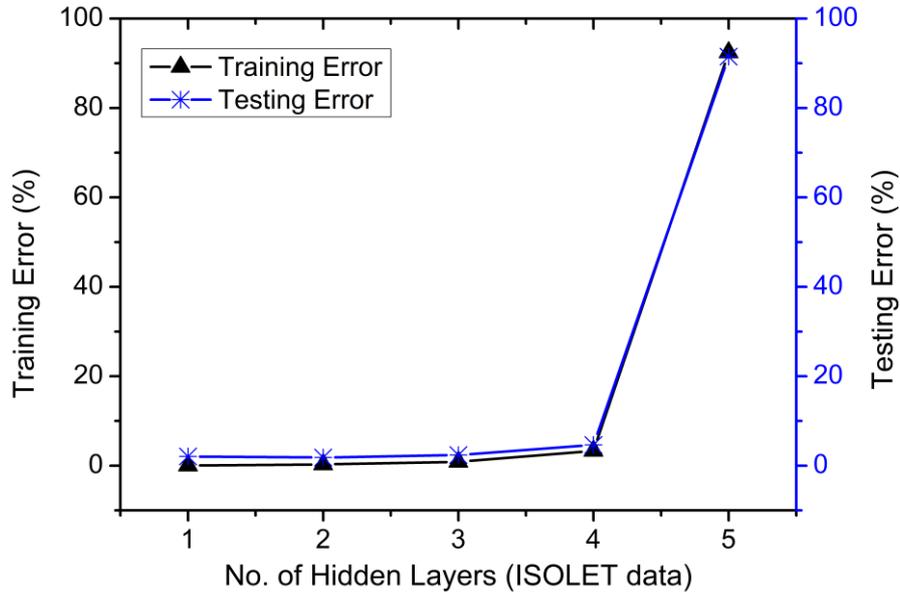

**d.** Performance of proposed algorithm on ISOLET dataset

**Fig. 2.** (a) – (d) Performance of the proposed algorithm in terms of training and testing error for different hidden layers.

# 6 Conclusion and Future Scope

This work shows that TS with GDM can be successfully applied to optimize deep FNN architecture. The solution is represented by hidden layers, hidden neurons and testing error, and considered as an entity of searching space. The aim of methodology is to find global optimal solution in which FNN has lowest testing error with good performance. Data set used in the experiment has large number of attributes and samples. The result generated show that the architecture for deep FNN can be suggested by proposed methodology. The interesting finding is that except for MNIST dataset, the proposed methodology suggested two hidden layer FNN architecture for each dataset for getting high accuracy. Hence, the proposed methodology can work in cases where FNN requires more than one hidden layer, i.e., can apply in Deep Neural Networks [30],[31].

As for future work, the work can be further extended to find optimal connections in the same solution. In this extension some more optimization techniques like SA, GA and PSO etc., can be combined to develop a hybrid approach based on the proposed methodology.

## Acknowledgement:

The author T.K. Gupta acknowledges Junior Research Fellowship (JRF) received from University Grant Commission (UGC), Ministry of Human Resources and Development, Govt. of India vide Reference No.: 3505/(NET-DEC.2013).